# RISK PRONENESS ESTIMATION METHOD DEVELOPED IN RELATION TO THE DECISION TAKER THAT CONTROLS THE ROBOTIC SYSTEM

V.Ya. Vilisov

*University of Technology, Energy IT LLP, Russia, Moscow Region, Korolev*
vvib@yandex.ru

**Abstract.** This work suggests the estimation method developed in relation to the position of the robotic system (RS) operator, showing his degree of risk proneness. The base models are: Hurwitz pessimism/optimism criterion and decision trees. The problem is solved using the reverse setting: we estimate pessimism/optimism parameter of the operator (decision taker) by observing what decisions he makes when controlling the RS. The solution context of such decision taker position estimation problems can be: using RS in emergency situations, in military actions and other situations connected with the uncertainty of the situation.

**Introduction.** Currently, operator-controlled robotic systems (RS) are widely used in a large application sphere such as emergency situations, space programs, military actions, medicine and some other spheres of RS application [1-4]. Everywhere in such applications the robot is a special manipulator providing the operator with more possibilities to influence the object of the manipulation. As a rule, all decisions in the situations occurring with RS are made by the operator who is actually a person that makes the decisions (decision taker) and the RS is actually his avatar, providing him with functionally new actuators and additional sensors that widen his basic capacities.

When applying RS in emergency situations (ES) or in conditions of active resistance (for example during the active military actions or space missions), the most important role is played by the time factor. It means that the operator always has to choose either he should spend some time on the situation update and analysis or he should act, using his experience and intuition, i.e. to use the RS to perform the assignment. Functions that assess the condition of the operating environment of the RS as well as some special operations are often performed by different specialized RS [5].

At that, the RS operators possess different level of experience, intuition, competence and risk proneness. Risk proneness position of the decision taker can range from extreme caution (pessimistic position) to high risk (optimistic position).

Current tendencies of RS development going in direction of their intellectualization [6] and higher degree of independence [7] make the tasks of providing RS with such abilities as significant as the presence of an effective operator is. In this connection we deem important the research [8, 9] that is aimed at construction of models referring to different aspects of controlling (managerial) activity performed by an effective operator for their further use in independent effective RS. This work considers one of aspects used for building such models, which is the estimation (identification) of the risk proneness degree (position) of the RS operator calculated by observing the decisions he makes.

**Problem Set-Up.** For the purposes of contextual certainty we shall take ES divisions that perform search & rescue and abandonment works at chemical or radioactive dangerous objects as an example. In such situations time is against the rescue workers, so that operators should make a choice on the basis of current situation: either they should perform additional situation appraisal or send the executor RS directly. After additional appraisal such dilemma appears again. A very cautious operator can perform additional appraisal for a long time not sending executor robots. A risky operator can send several robots at once to perform the task, and the further inspection will either prove him right or wrong. As a rule, experience and competence of the operator being the decision taker should provide maximum effectiveness of rescue operations. When the operator makes a decision, the degree of his risk proneness can be justified (i.e. providing an acceptable effectiveness to the whole rescue operation) or non-justified. This gives us grounds to suggest that for some certain ES there is an acceptable (allowable, effective) risk degree when making decisions.

Thus we have a problem: what indicator should be used to measure the operator's proneness to risk and how it should be estimated when observing the decisions that are made by a specific decision taker. The formal characterization of the problem is going to be executed by using the decision trees (nature games, positional strategic games) [10, 11] and Hurwitz pessimism/optimism criterion [12], which uses the parameter (indicator) that shows the decision taker's degree of risk proneness when making decisions. The formal presentation of the suggested method shall be performed using the model example.

The operator's (decision taker) decision making problem shall be presented as a three-level decision tree (Picture 1) where the first (the lowest on the tree) outcome level $(a, b, c, d)$ corresponds to four levels of the state of operating environment when there appeared a necessity in making a decision ($a$ - absence of works for executor RS; $b, c, d$ - three levels of work scope in ascending order). However, these states possessing a different degree of certainty cannot be a priori known to the operator. They can be presented, for example, as a distribution of possibilities at the multitude of discrete states. Thus, the degree of certainty may vary from complete uncertainty $(P(a) = P(b) = P(c) = P(d) = 0.25)$ to complete certainty, for example $P(a) = P(b) = P(c) = 0; P(d) = 1$. Further it is presumed that the decision taker cannot assess the degree of uncertainty of a current state, only feeling it intuitively, thus making a decision regarding sending (1) or not sending (0) executor RS to execute the assignments (see the second level of the decision tree).

At the third level the outcomes $(a, b, c, d)$ have the same meaning as they have at the first one, but here some certain state (which is found by the executor RS) brings the specific values of the execution indicators (see the figures at the top of the tree) such as an area of the extinguished fire or a volume of the resource that was used to execute the assignment etc. Without loss of generality, let us assume that the outcome indicators located on the decision tree have the meaning of payments that the operator tries to maximize.

If, when making decisions, the decision taker does not use any formal constructions but makes his choice on the basis of personal experience and intuition, then, in proportion to his accumulated experience, considering many factors and his own hierarchy of values that were obtained and changed due to his own understanding of things, the alternatives chosen by him will become more and more effective. At that let us assume that there is a reverse connection existing, for example, in the form of posterior estimation of effectiveness of his actions taken when making decisions in relation to the situation or within some period of time. Such estimation of effectiveness can be performed by the senior control level, by an authorized person or collectively by some group of people.

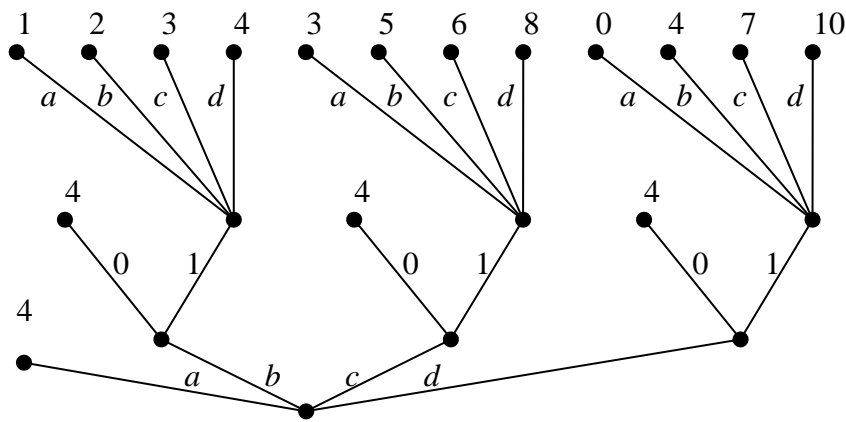

Pic. 1. RS Control Decision Tree

**Solution.** Let us consider the totality of the decision choice criteria presented by the combined Hurwitz criterion, on the basis of which we shall build the procedure of discovering the decision taker's position. The observations that we use herein are the "good" (effective) decisions used for building the choice model presented by the decision tree (see Picture 1). Within the multistep procedures of decision making, the algorithm of decision taker position identification shall be presented as a following sequence of stages.

Stage 1. To execute the decision tree normalization procedure (presenting the problem in the normal form: as a matrix or a table) using one of existing methods [12]. At that the payments shall be presented as a payment matrix $A = \|a_{hj}\|_{mn}$ where $m$ is a value of pure strategies of the decision taker (lines) and $n$ is a value of nature conditions (columns).

Stage 2. To build the dependency (from $\lambda$ parameter) of the good strategy $f(\lambda)$ that was obtained using the Hurwitz criterion:

$$V = \max_i L(i) = \max_i \left( \lambda \min_j a_{ij} + (1 - \lambda) \max_j a_{ij} \right), \qquad (1)$$

for example, by changing $\lambda$ parameter at some regular grid within [0;1] interval. As a result of such calculations, we shall build the dependency $f(\lambda)$, with the help of which the reverse $\lambda(f)$ shall be built.

Stage 3. Using the statistical data gathered when observing "good" decisions of the decision taker, i.e. using the most probable strategy $f$ out of all he would apply, and on the basis of reverse dependency $\lambda(f)$ we shall compute the $\lambda$ parameter that will correspond to the decision taker's position.

Let us consider the realization of the algorithm taking into account model data presented at Picture 1.

For applying the algorithm of decision taker position identification let us generate implementations of the decisions (Table 1) made by the decision taker in the imitation mode for the following probabilities of nature conditions at the first and third steps: $P(a) = 0.3$; $P(b) = 0.3$; $P(c) = 0.3$; $P(d) = 0.1$. At that we shall assume that the decision taker possesses adequate experience, which allows us to consider all his decisions as "good" ones; thus they can be used in the statistical estimation. In the imitation mode we shall set $\lambda = 0.7$, i.e. the decision taker's position is close to cautious (here it is necessary to mention that $\lambda = 1$ corresponds to the position of extreme caution and extreme pessimism, while $\lambda = 0$ corresponds to the position of extreme optimism). Table 1 contains a segment of the simulated condition observation sample (conditions of nature, operating environment), decisions made by the decision taker and final payment (effect).

Table 1. Segment of Observation Sample

| Observations | 1 | 2 | 3 | 4 | 5 | 6 | 7 | 8 | 9 | 10 | 11 | 12 | 13 | 14 | 15 |
|---|---|---|---|---|---|---|---|---|---|---|---|---|---|---|---|
| Step 1 (Nature) | b | b | d | c | b | b | b | b | c | b | b | c | d | c | d |
| Step 2 (Decision Taker) | 0 | 0 | 0 | 1 | 0 | 0 | 0 | 0 | 1 | 0 | 0 | 1 | 0 | 1 | 0 |
| Step 3 (Nature) | c | a | d | d | c | c | a | b | a | c | b | a | c | d | b |
| Payment | 4 | 4 | 4 | 8 | 4 | 4 | 4 | 4 | 3 | 4 | 4 | 3 | 4 | 8 | 4 |

After having understood the state of the nature at the first step of the tree, it is possible to present the action of the operator as a strategy vector:

$$f = \begin{bmatrix} i \\ j \\ k \end{bmatrix}, \qquad i, j, k \in \{0; 1\}, \qquad (2)$$

where 0 and 1 are the alternatives, basing on which the decision taker makes a choice (see Picture 1 and Table 1) provided that at the first step the state was $b, c, d$ correspondingly. In this case eight strategies are possible, and the RS operator uses one of them:

$$f_1 = \begin{bmatrix} 0 \\ 0 \\ 0 \end{bmatrix}; f_2 = \begin{bmatrix} 0 \\ 0 \\ 1 \end{bmatrix}; f_3 = \begin{bmatrix} 0 \\ 1 \\ 0 \end{bmatrix}; f_4 = \begin{bmatrix} 0 \\ 1 \\ 1 \end{bmatrix}; f_5 = \begin{bmatrix} 1 \\ 0 \\ 0 \end{bmatrix}; f_6 = \begin{bmatrix} 1 \\ 0 \\ 1 \end{bmatrix}; f_7 = \begin{bmatrix} 1 \\ 1 \\ 0 \end{bmatrix}; f_8 = \begin{bmatrix} 1 \\ 1 \\ 1 \end{bmatrix}. \qquad (3)$$

Using the implementation shown in Table 1 we can see that if at the first step the nature is in $b$ or in $d$ condition, the decision taker chooses the 0 alternative; and if the nature is $c$, he chooses the 1 alternative, which means that he uses $f_3$ strategy.

Further we shall execute the stage-by-stage algorithm provided above.

Stage 1. Here we shall execute the decision tree normalization procedure. For this we have to determine the set elements of pure strategies of the decision taker and set elements of the nature conditions. The decision taker strategies have already been defined by the correlation (3). The multitude of nature conditions shall be defined as a multitude of combinations pertaining to outcomes of the first and the third steps, excluding the outcome $a$ at the first step as here the decision taker does not make any decisions. Thus, nature can have one out of 12 states with each of them presented as a vector of possible states of the operating environment at the first and the second steps of the decision tree correspondingly: $s_j = [x \quad y]^T$ where $T$ is a conjugation symbol; $x \in \{b, c, d\}$; $y \in \{a, b, c, d\}$. Then the whole set of states will be the following:

$$s_1 = \begin{bmatrix} b \\ a \end{bmatrix}; \; s_2 = \begin{bmatrix} b \\ b \end{bmatrix}; \; s_3 = \begin{bmatrix} b \\ c \end{bmatrix}; \; s_4 = \begin{bmatrix} b \\ d \end{bmatrix}; \; s_5 = \begin{bmatrix} c \\ a \end{bmatrix}; \; s_6 = \begin{bmatrix} c \\ b \end{bmatrix};$$
$$s_7 = \begin{bmatrix} c \\ c \end{bmatrix}; \; s_8 = \begin{bmatrix} c \\ d \end{bmatrix}; \; s_9 = \begin{bmatrix} d \\ a \end{bmatrix}; \; s_{10} = \begin{bmatrix} d \\ b \end{bmatrix}; \; s_{11} = \begin{bmatrix} d \\ c \end{bmatrix}; \; s_{12} = \begin{bmatrix} d \\ d \end{bmatrix}.$$
(4)

Let us show the payments that correspond to the outcomes of the decision tree as Table 2, where, for purposes of brevity, we shall present the states using only a pair of their coordinate values.

Table 2. Normalized Payment Matrix

|  |  | $s_j$ | | | | | | | | | | | |
|---|---|---|---|---|---|---|---|---|---|---|---|---|---|
|  |  | ba | bb | bc | bd | ca | cb | cc | cd | da | db | dc | dd |
| $f_h$ | 000 | 4 | 4 | 4 | 4 | 4 | 4 | 4 | 4 | 4 | 4 | 4 | 4 |
|  | 001 | 3 | 3 | 3 | 3 | 3 | 3 | 3 | 3 | 0 | 4 | 7 | 10 |
|  | 010 | 4 | 4 | 4 | 4 | 3 | 5 | 6 | 8 | 4 | 4 | 4 | 4 |
|  | 011 | 4 | 4 | 4 | 4 | 3 | 5 | 6 | 8 | 0 | 4 | 7 | 10 |
|  | 100 | 1 | 2 | 3 | 4 | 4 | 4 | 4 | 4 | 4 | 4 | 4 | 4 |
|  | 101 | 1 | 2 | 3 | 4 | 4 | 4 | 4 | 4 | 0 | 4 | 7 | 10 |
|  | 110 | 1 | 2 | 3 | 4 | 3 | 5 | 6 | 8 | 4 | 4 | 4 | 4 |
|  | 111 | 1 | 2 | 3 | 4 | 3 | 5 | 6 | 8 | 0 | 4 | 7 | 10 |

Stage 2. In order to build the dependency of $f$ good strategy from the parameter $\lambda$ of the Hurwitz criterion (1), we shall vary it with the 0.1 step at the interval [0; 1], computing the values of objective function $L(i)$ for each step. The computational results are provided in Table 3.

The last line in the table is the table record of $f(\lambda)$ function, where we can see that the reverse dependency $\lambda(f)$ is a multidigit one, i.e. one value of the argument corresponds to the interval of values, for example $f_1$ corresponds to the value interval of $\lambda$ ranging from 0.8 to 1.0.

Stage 3. On the basis of the statistical data gathered from the alternative choice observations of the decision taker (see Table 1) we can deduce that the decision taker follows the $f_3$ strategy. And the table dependency $\lambda(f)$ shows that the coefficient of the Hurwitz criterion that reflects the decision taker's position is located within [0.5; 0.7] interval which complies with the original modelling data (we have set the $\lambda = 0.7$ value). The reverse problem has been solved.

Table 3. Objective Function Values According to the Hurwitz Criterion

|  |  | $\lambda$ | | | | | | | | | | |
|---|---|---|---|---|---|---|---|---|---|---|---|---|
|  |  | 0 | 0.1 | 0.2 | 0.3 | 0.4 | 0.5 | 0.6 | **0.7** | 0.8 | 0.9 | 1.0 |
| $f_h$ | 000 | 4.0 | 4.0 | 4.0 | 4.0 | 4.0 | 4.0 | 4.0 | 4.0 | **4.0** | **4.0** | **4.0** |
|  | 001 | **10.0** | **9.0** | **8.0** | **7.0** | **6.0** | 5.0 | 4.0 | 3.0 | 2.0 | 1.0 | 0.0 |
|  | **010** | 8.0 | 7.5 | 7.0 | 6.5 | **6.0** | **5.5** | **5.0** | **4.5** | **4.0** | 3.5 | 3.0 |
|  | 011 | **10.0** | **9.0** | **8.0** | **7.0** | **6.0** | 5.0 | 4.0 | 3.0 | 2.0 | 1.0 | 0.0 |
|  | 100 | 4.0 | 3.7 | 3.4 | 3.1 | 2.8 | 2.5 | 2.2 | 1.9 | 1.6 | 1.3 | 1.0 |
|  | 101 | **10.0** | **9.0** | **8.0** | **7.0** | **6.0** | 5.0 | 4.0 | 3.0 | 2.0 | 1.0 | 0.0 |
|  | 110 | 8.0 | 7.3 | 6.6 | 5.9 | 5.2 | 4.5 | 3.8 | 3.1 | 2.4 | 1.7 | 1.0 |
|  | 111 | **10.0** | **9.0** | **8.0** | **7.0** | **6.0** | 5.0 | 4.0 | 3.0 | 2.0 | 1.0 | 0.0 |
| $L^*(\lambda)$ |  | 10.0 | 9.0 | 8.0 | 7.0 | 6.0 | 5.5 | 5.0 | 4.5 | 4.0 | 4.0 | 4.0 |
| $f^*(\lambda)$ |  | $f_2$ | $f_2$ | $f_2$ | $f_2$ | $f_2$ | $f_3$ | $f_3$ | $f_3$ | $f_1$ | $f_1$ | $f_1$ |

**Conclusions.** The results obtained from the data of the model example can be explained in such a manner that having presented with the reiterated choice of decisions, the specific decision taker will receive a five unit payment on average by keeping to his $f_3$ strategy. At that on the basis of the reverse problem solution it is possible to conclude that his risk proneness indicator is within [0.5; 0.7] interval. If such control-obtained effect is deemed satisfactory, in the future such level of risk proneness can be considered adequate; using it as a base, RS can be equipped with the systems of decision making support applied in similar situations and/or the suggested decision making algorithm can be included in the software of the corresponding RS control system level.

Thus, the suggested approach allows to provide the risk level that is considered adequate and which was approved by the experts who evaluate quality and effectiveness of the operations performed with the use of robotic systems.

**References**

[1] N.A. Rudianov, A.V. Ryabov, V.S. Khruschov Land Robotic Systems Being the Defense Element of Objects and Territory of the Russian Federation // Book of reports of the Russian National Scientific and Technological Conference "Extreme Robotics", Saint Petersburg: Politekhnika-Service, 2015, pp. 16-17.

[2] S.G. Tsarichenko, A.V. Ivanov, Yu.N. Osipov, A.Yu. Kartenichev, V.I. Yershov Peculiarities of UAS Application by the Ministry of Emergencies // Book of reports of the Russian National Scientific and Technological Conference "Extreme Robotics", Saint Petersburg: Politekhnika-Service, 2015, pp. 24-29.

[3] S.R. Lysyi Scientific and Technological Problems and Perspectives of Development of Special (Space) Robotics // Book of reports of the Russian National Scientific and Technological Conference "Extreme Robotics", Saint Petersburg: Politekhnika-Service, 2015, pp. 29-34.

[4] K.Ye. Senchik, V.V. Kharlamov, N.A. Gryaznov, A.V. Lopota On Peculiarities of Robotics Application in Medicine // Book of reports of the Russian National Scientific and Technological Conference "Extreme Robotics", Saint Petersburg: Politekhnika-Service, 2015, pp. 40-43.

[5] I.A. Kalyayev, S.G. Kapustyan Problems of Robotic Group Control // Mechatronics, Automation, Control, No. 6, 2009, pp. 33-40.

[6] M.V. Lokhin, S.V. Manko, M.P. Romanov Increasing Adaptive Functions of Autonomous Robots on the Base of Intellectual Technologies // Book of reports of the Russian National Scientific and Technological Conference "Extreme Robotics", Saint Petersburg: Politekhnika-Service, 2015, pp. 63-67.

[7] A.A. Zhdanov Independent Artificial Intellect. - Moscow: BINOM, 2008, p. 359.

[8] V.Ya. Vilisov On Robot Algorithms Adapted to the Target Preferences of the Decision Taker // Book of reports of the Russian National Scientific and Technological Conference "Extreme Robotics", Saint Petersburg: Politekhnika-Service, 2012, pp. 120-126.

[9] V.Ya. Vilisov Adaptive Choice of Managerial Decisions. Operation Examination Models as the Means of Storing the Decision Taker's Knowledge - Saarbruecken (Germany): LAP LAMBERT Academic Publishing, 2011, p. 376.

[10] G. Raifa Decision Analysis. - Moscow: Nauka, 1977, p. 408.

[11] Guillermo Owen Game Theory. - Moscow: Mir, 1971. - p. 230.

[12] Hamdy A. Taha Introduction to Operations Research: Translation from English - Moscow: Williams Publishing House, 2005, p. 912.